\documentclass[11pt,a4paper]{article}
\usepackage[hyperref]{emnlp2020}
\usepackage{times}
\usepackage{latexsym}
\usepackage{amssymb}
\usepackage{amsmath}
\usepackage{amsfonts}
\usepackage{xspace}
\usepackage{algorithm}
\usepackage{algorithmic}
\usepackage{tabularx}
\usepackage{soul}
\usepackage{booktabs}
\usepackage{appendix}
\usepackage{colortbl}
\usepackage{arydshln}
\usepackage{multirow}
\usepackage{multicol}
\usepackage{paralist}
\usepackage{array}
\usepackage{hyperref}
\usepackage{adjustbox}

\usepackage{microtype}

\aclfinalcopy 
\def\aclpaperid{3205} 

\newcommand{\name}{CSS\xspace}
\newcommand{\DSS}{PSS\xspace}
\newcommand{\supe}{Sup\xspace}
\newcommand{\unsup}{UnSup\xspace}
\newcommand{\POT}{POT\xspace}
\newcommand{\BLU}{BLU\xspace}

\newcommand{\real}{\ensuremath{\mathbb{R}}}

\newcommand{\tabincell}[2]{\begin{tabular}{@{}#1@{}}#2\end{tabular}} 
\newcolumntype{C}{ >{\centering\arraybackslash} m{0.5cm} }
\newcolumntype{B}{ >{\arraybackslash} m{0.2cm} }
\newcolumntype{D}{ >{\arraybackslash} m{3.5cm} }
\newcolumntype{E}{ >{\arraybackslash} m{2.8cm} }
\newcolumntype{F}{ >{\arraybackslash} m{2cm} }
\newcolumntype{X}{ >{\arraybackslash} m{1.9cm} }
\newcolumntype{Y}{ >{\arraybackslash} m{0.27cm} }
\makeatletter
\def\hlinew#1{%
  \noalign{\ifnum0=`}\fi\hrule \@height #1 \futurelet
   \reserved@a\@xhline}

\usepackage{todonotes}
\usepackage{soul}
\soulregister\cite7 
\soulregister\citep7 
\soulregister\citet7 
\soulregister\ref7 
\soulregister\pageref7 
\soulregister\name7

\title{Semi-Supervised Bilingual Lexicon Induction \\ with Two-way Interaction}

\author{Xu Zhao$^1$, Zihao Wang$^2$, Hao Wu$^3$, Zhang Yong$^1$ \footnotemark[2] \\
  $^1$BNRist, Department of Computer Science and Technology, RIIT,\\
	Institute of Internet Industry, Tsinghua University, Beijing, China \\
  $^2$Department of CSE, HKUST, Hong Kong SAR, China\\
  $^3$Department of Mathematical Sciences, Tsinghua University, Beijing, China\\
  \texttt{zhaoxu18@mails.tsinghua.edu.cn} \\
  \texttt{zwangggc@cse.ust.hk} \\
  \texttt{\{hwu,zhangyong05\}@tsinghua.edu.cn}}

\date{}

\begin{document}
\maketitle
\renewcommand{\thefootnote}{\fnsymbol{footnote}}
\footnotetext[2]{Yong Zhang is the corresponding author.}
\begin{abstract}
Semi-supervision is a promising paradigm for Bilingual Lexicon Induction (BLI) with limited annotations.
However, previous semi-supervised methods do not fully utilize the knowledge hidden in annotated and non-annotated data, which hinders further improvement of their performance.
In this paper, we propose a new semi-supervised BLI framework to encourage the interaction between the supervised signal and unsupervised alignment. We design two message-passing mechanisms to transfer knowledge between annotated and non-annotated data, named prior optimal transport and bi-directional lexicon update respectively. Then, we perform semi-supervised learning based on a cyclic or a parallel parameter feeding routine to update our models. Our framework is a general framework that can incorporate any supervised and unsupervised BLI methods based on optimal transport. Experimental results on MUSE and VecMap datasets show significant improvement of our models. Ablation study also proves that the two-way interaction between the supervised signal and unsupervised alignment accounts for the gain of the overall performance. Results on distant language pairs further illustrate the advantage and robustness of our proposed method.

\end{abstract}

\section{Introduction}
Bilingual Lexicon Induction (BLI) is of huge interest to the research frontier. BLI methods learn cross-lingual word embeddings from separately trained monolingual embeddings. BLI is believed to be a promising way to transfer semantic information between different languages, and spawns lots of NLP applications like machine translation \citep{DBLP:conf/emnlp/LampleOCDR18, DBLP:conf/iclr/ArtetxeLAC18}, Part Of Speech (POS) tagging \citep{gaddy2016ten}, parsing \citep{DBLP:conf/conll/XiaoG14}, and document classification \citep{DBLP:conf/coling/KlementievTB12}.

The key step of BLI is to learn a transformation between monolingual word embedding spaces~\citep{DBLP:journals/jair/RuderVS19}, which could be further used for translation retrieval or cross-lingual analogy tasks.
However, it is hard to obtain the high quality transformation with low supervision signals, i.e. with limited annotated lexicon.
Thus, some semi-supervised BLI methods~\citep{DBLP:conf/acl/ArtetxeLA17, DBLP:conf/acl/PatraMGGN19} are proposed to make use of annotated and non-annotated data. \citet{DBLP:conf/acl/ArtetxeLA17} bootstrapped the supervised lexicon to enhance the supervision but ignored the knowledge in non-annotated data.
Meanwhile, \citet{DBLP:conf/acl/PatraMGGN19} combined the unsupervised BLI loss that captured the structural similarity in word embeddings \citep{DBLP:conf/iclr/LampleCRDJ18} with the supervised loss \citep{DBLP:conf/emnlp/JoulinBMJG18}.
However, this loss combination still performed poorly since the bad supervised optimization under limited annotations, see the Experiment part for details.
As a result, existing semi-supervised BLI methods suffer from low effectiveness~\citep{DBLP:conf/acl/ArtetxeLA17} or low robustness~\citep{DBLP:conf/acl/PatraMGGN19}.

\begin{figure*}
    \centering
    \includegraphics[width=.7\linewidth]{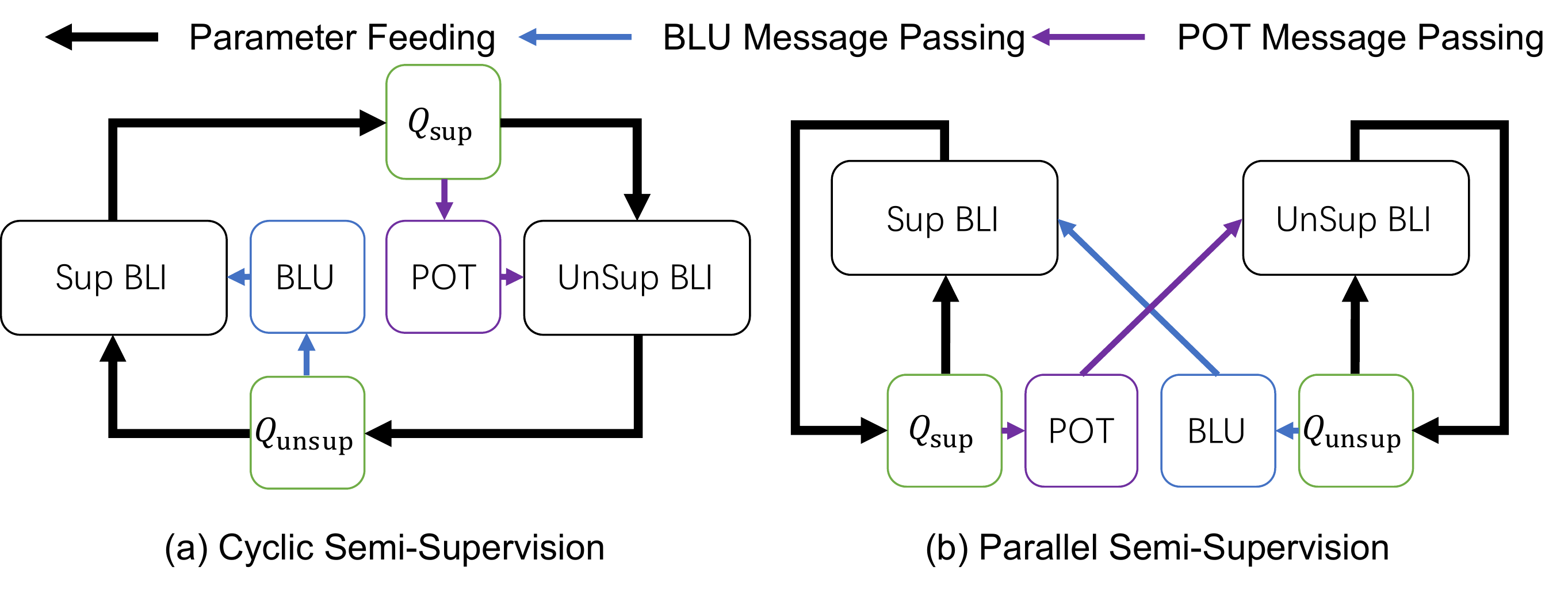}
    \caption{Illustration for Cyclic Semi-Supervision and Parallel Semi-Supervision}
    \label{fig:css-dss}
\end{figure*}

In this work, we focus on designing a new semi-supervised BLI method to make full use of both annotated and non-annotated data.
We propose a novel framework with two different strategies, which exceeds the previous separate~\citep{DBLP:conf/acl/ArtetxeLA17,DBLP:conf/acl/PatraMGGN19} semi-supervised methods by emphasizing the two-way interaction between the supervised signal and unsupervised alignment.
In this framework, supervised training tries to align the parallel lexicon and unsupervised training can exploit the structure similarity between monolingual embedding spaces. 
The foundation of two-way interaction is in two carefully designed \emph{message passing mechanisms}, see Section~\ref{sec:prior} and~\ref{sec:LU}.
Two-way interaction enables semi-supervised BLI to guide the exploitation of structural similarity by \emph{unsupervised procedure}~\citep{DBLP:conf/aistats/GraveJB19} and extend insufficient lexicon for \emph{supervised procedure}~\citep{DBLP:conf/emnlp/JoulinBMJG18} simultaneously, see Figure~\ref{fig:css-dss}. In this paper, we only consider the unsupervised BLI methods based on Optimal Transport (OT)~\citep{DBLP:conf/aistats/GraveJB19, DBLP:journals/corr/abs-1811-01124,DBLP:conf/acl/HuangQC19,DBLP:conf/aistats/Alvarez-MelisJJ18}, which have achieved impressive results on BLI task. 



More specifically, the contributions of this paper are listed below.
\begin{compactitem}
\item We propose the two-way interaction between the supervised signal and unsupervised alignment. It consists of two message passing mechanisms, Prior Optimal Transport~(\POT) and Bidirectional Lexicon Update (\BLU). \POT enables the OT-based unsupervised BLI approach to be guided by any prior BLI transformation, i.e. transfers what is learned by supervised BLI method to the unsupervised BLI method.
\BLU employs the alignment results in bi-directional retrieval to enlarge the annotated data, and thus enhances the supervised training by unsupervised BLI transformation.
\item We propose two strategies of semi-supervised BLI framework based on \POT and \BLU, named by Cyclic Semi-Supervision (\name) and Parallel Semi-Supervision (\DSS).
They are recognized by the cyclic and parallel parameter feeding routines, respectively, see Figure~\ref{fig:css-dss}. Notably, \name and \DSS are universal to admit any supervised BLI methods and OT-based unsupervised BLI methods.
\item Extensive experiments on two popular datasets show that \name and \DSS exceed all previous supervised, unsupervised, and semi-supervised approaches and are suitable to different scenarios. Ablation study of \name and \DSS demonstrates that the two-way interaction (\POT and \BLU) is the key to improve the performance. Results on distant language pairs show the advantage and robustness of our method.
\end{compactitem}

\section{Background}~\label{sec:background}
In this section, we describe the basic formulation of related supervised and unsupervised BLI methods. We define two embedding matrices $X, Y \in \mathbb{R}^{n \times d}$,  where $n$ is the number of words and $d$ is the dimension of the word embedding. 

The key to supervised BLI is the parallel lexicon between two languages, say word $x_i$ in $X$ is translated to word $y_i$ in $Y$. \citet{DBLP:journals/corr/MikolovLS13} suggested regarding supervised BLI as a regression problem aligning word embeddings by a linear transformation $Q^{\star}$.
\begin{equation} \label{eq:procrutes}
Q^\star = \mathop{\arg\min}_{Q \in \mathbb{R}^{d\times d}} \sum_i \|X_i Q-Y_i\|^{2}_{2}
\end{equation}
\citet{DBLP:conf/emnlp/ArtetxeLA16} introduced the orthogonal constraint on $Q$. Therefore, Problem~\eqref{eq:procrutes} has a closed-form solution $Q^{\star}=UV^\top$, where $U, V$ are defined by the SVD decomposition $Y^\top X = USV^\top$.
\citet{DBLP:conf/emnlp/JoulinBMJG18} proposed to replace the $2$-norm in Problem~\eqref{eq:procrutes} by Relaxed Cross-domain Similarity Local Scaling (RCSLS) loss to mitigate the hubness problem, which is formulated in Equation~\eqref{eq:rcsls}.
\begin{equation}
	\begin{split}
		\texttt{RCSLS}(x_i Q, y_i) =& -2 Q^{\top}x_i^{\top}  y_i 
		\\ + &\frac{1}{k} \sum_{y \in \mathcal{N}_{Y}\left(x_i Q\right)} Q^\top x_i^\top y
		\\ + &\frac{1}{k} \sum_{xQ \in \mathcal{N}_{X}\left(y_j\right)} Q^\top x^\top y_j
	\end{split}\label{eq:rcsls}
\end{equation}
where $\mathcal{N}_{X}\left(y\right)$ represents the set which consists of the $k$ nearest neighbors of $y$ in the point cloud $X$, so as $\mathcal{N}_{Y}\left(x_iQ\right)$.


For unsupervised BLI, embeddings in $X$ and $Y$ are totally out of order. As a result, unsupervised BLI methods need to model an unknown permutation matrix $P\in \mathcal{P}_n = \{0, 1\}^{n\times n}$
\begin{equation}
\min_{Q \in \mathcal{O}_d, P \in \mathcal{P}_n} \|XQ - PY\|^{2}_{F}
\label{eq:unsupervised-proc}
\end{equation}
where $\mathcal{O}_{d}$ is the set of orthogonal matrices. Problem~\eqref{eq:unsupervised-proc} could be solved by iteratively minimizing $Q$ and $P$.
More specifically, \citet{DBLP:conf/aistats/GraveJB19} considered random samples $\bar X, \bar Y \in \mathbb{R}^{m\times d}$ from $X, Y$ in a stochastic optimization scheme.
Minimizing $P$ directly is hard.
The key to unsupervised methods is how to solve $P$ approximately, see Section \ref{sec:relatedwork}.
OT based methods solve $P$ by optimal transport~\citep{zhang2017earth, DBLP:conf/aistats/GraveJB19, DBLP:journals/corr/abs-1811-01124, DBLP:conf/acl/HuangQC19}.
\citet{DBLP:conf/aistats/GraveJB19} and \citet{zhang2017earth} proposed to solve the Wasserstein problem between the two distributions supported on $\bar{X}Q$ and $\bar{Y}$, respectively.
\begin{equation}
	W^2_2(\bar{X}Q, \bar{Y}) = \min_{P \in \Pi}  \sum_{i, j} P_{ij} D_{ij}
\label{eq:wasserstein-proc}
\end{equation}
where $D_{ij}$ is the cost between $x_i Q$ and $y_j$, such as $2$-norm, RCSLS loss, or other costs.
$P \in \Pi = \{P\in \mathbb{R}^{m\times m}_+ | \sum_i P_{ij} =\sum_j P_{ij} = 1\}$ is the transport plan~\cite{peyre2019computational}.
OT related metrics can be solved by the entropy regularized Sinkhorn algorithm~\citep{DBLP:conf/nips/Cuturi13}.
\begin{equation}
	W^2_2(\bar{X}Q, \bar{Y}) = \min_{P \in \Pi}  \sum_{i, j} P_{ij} D_{ij} + \epsilon H(P)
\label{eq:entropic-wasserstein-proc}
\end{equation}

To summarize, the foundation of supervised BLI is the annotated parallel lexicon for training, and the critical step of OT-based unsupervised BLI is the solution of transport plan $P$.

\section{Message Passing in BLI}
In this section, we present two message passing mechanisms for semi-supervised framework, including \POT and \BLU.
\POT is proposed to enhance the unsupervised BLI by the knowledge passed from the supervised BLI.
Meanwhile \BLU enhances the supervised BLI by the additional lexicon based on the unsupervised retrieval results.
Therefore, \POT and \BLU form the two-way interaction between the supervised signal and unsupervised alignment.

\subsection{Prior Optimal Transport}~\label{sec:prior}
We present \POT to strengthen the stochastic optimization in unsupervised BLI with prior information from supervised BLI.
More specifically,
\POT is designed to guide the original OT solution of $P$, see Problem~\eqref{eq:wasserstein-proc}.
\POT can replace the original OT problems in unsupervised BLI models such as~\cite{DBLP:conf/aistats/GraveJB19}.
In this way, we enable the transformation $Q_{sup}$ trained in supervised BLI to enhance the unsupervised BLI.
 
Given $Q_{sup}$ learned from any supervised BLI and random word embedding samples $\{x_i\}$ and $\{y_j\}$, we compute the cost matrix $C$ between transformed source embeddings and target embeddings. In this work, we choose RCSLS as the specific formulation of $C_{ij}$
 \begin{equation}
 	C_{ij} = RCSLS(x_i Q_{sup}, y_j).
 \end{equation}

Based on this cost function, we propose to use the Boltzmann Distribution, i.e. softmax function with temperature to construct a prior transport plan $\Gamma$:
\begin{equation} \label{eq:prior-distribution}
	\Gamma_{ij}= \frac{e^{- C_{ij}/T}}{\sum_{1 \leq k \leq n} e^{-C_{ik}/T}}.
\end{equation}
$\Gamma_{ij}$ represents the probability that the $i$-th word in $X$ is a translation of the $j$-th word in $Y$. Temperature $T$ controls the significance of translation in $\Gamma$. 
$\Gamma$, induced from $Q_{sup}$, assigns each pair of words with a smaller cost in $C$ a higher probability of forming a lexicon.

Instead of considering Problem~\eqref{eq:wasserstein-proc}, we consider the \POT regularized by the Kullback-Leibler (KL) divergence between $\Gamma$ and $P$.
\begin{align} \label{eq:prior-ot}
    \texttt{POT}(\bar XQ, \bar Y) = \min_{P \in \Pi}  \langle D, P  \rangle + \varepsilon KL(P\|\Gamma),
\end{align}
where $ \langle D, P  \rangle = \sum_{i, j} P_{ij} D_{ij}$ is the matrix inner product.
We note that KL regularization in \POT problem is totally different from the aforementioned entropic regularization~\eqref{eq:entropic-wasserstein-proc}.
For entropy regularized OT, the regularization coefficient $\epsilon$ is expected to be as small as possible to approximate the original OT solution.
However, for \POT discussed in~\eqref{eq:prior-ot}, the regularization coefficient $\varepsilon$ controls the interpolation of OT transport plan that minimizes Problem~\eqref{eq:wasserstein-proc} and prior transport plan $\Gamma$.
Therefore, $\varepsilon$ does not need to be as small as possible.
Instead, it is a proper number to coordinate the effect from prior supervised transformation $Q_{sup}$.

The key to solving Problem~\eqref{eq:prior-ot} is to decompose the KL divergence into entropic term and linear term.
Therefore, Problem~\eqref{eq:prior-ot} is reduced to
\begin{align}
    \min_{P \in \Pi}  \langle D - \varepsilon \log \Gamma , P  \rangle + \varepsilon H(P).
\end{align}

By treating $D - \varepsilon \log \Gamma$ as the $\Gamma$-prior cost matrix $D(\varepsilon, \Gamma)$, Problem~\eqref{eq:prior-ot} could also be solved by Sinkhorn algorithm.
Again, since \POT does not require $\varepsilon$ to be closed to zero, the solution of \POT Problem~\eqref{eq:prior-ot} will not suffer from numerical instability problems~\citep{peyre2019computational}.

\subsection{Bi-directional Lexicon Update} \label{sec:LU}
As stated in Section~\ref{sec:background}, the key to supervised BLI is its parallel lexicon for training.
Therefore, to enhance the supervised BLI, we propose \BLU to extend the parallel lexicon by the structural similarity of word embeddings exploited in unsupervised BLI.
To distinguish from unsupervised notations in Section~\ref{sec:prior}, let $S, T \in \real^{n \times d}$ be the parallel word embedding matrices for source and target languages respectively.
The $i$-th row $s_i$ of $S$ and $t_i$ of $T$ form a translation pair in the annotated lexicon.
\BLU selects the additional lexicon $S', T'\in \real^{l\times d}$ with high credit scores to extend $S$ and $T$. Let $S^* = S \oplus S' $ and $T^* = T\oplus T' $ be the extended lexicon, where $\oplus$ denotes the concatenation operation along columns between two matrices.

Given the forward and backward transformations $\overrightarrow{Q}_{unsup}$ and $\overleftarrow{Q}_{unsup}$ between source language and target language from unsupervised BLI.
\BLU defines the $S'$ and $T'$ by $Q_{unsup}$ and $Q_{unsup}^\top$ respectively in four steps:

\noindent\textbf{(1) Compute the forward and backward distance matrices.}
Forward distance matrix $\overrightarrow{D}$ is defined between transformed source embeddings and target embeddings, while backward distance matrix $\overleftarrow{D}$ is defined between source embeddings and backward transformed target embeddings.

\noindent\textbf{(2) Generate forward and backward translation pairs.} 
Let $\overrightarrow{B} = \{(i,j)| j = \arg\min_k \overleftarrow{D}_{ik} \}$ and $\overleftarrow{B} = \{(i,j)| i = \arg\min_k \overleftarrow{D}_{kj} \}$ be the translation pair sets. Then take the intersection  $B = \overrightarrow{B} \cap \overleftarrow{B}$ as the candidate additional lexicon.

\noindent\textbf{(3) Compute the credit score $CS$ for each translation pair.}
Firstly, we define the forward and backward credit scores for a pair $(i,j) \in B$.
Let $\overrightarrow{C}(i)$ be the set of target word indices k, $\overrightarrow{C}(i) = $\{$k|$ $k \neq j$ and k is among top $K + 1$ elements of $-\overrightarrow{D}_{i*}$\}, so as $\overleftarrow{C}(j)$. The forward credit score is defined by $\overrightarrow{CS}_{ij} = \sum_{k\in \overrightarrow{C}(i)} \overrightarrow{D_{ik}}/K - \overrightarrow{D}_{ij}$, and $\overleftarrow{CS}_{ij}$ is similarly defined.
Then we define credit score $CS_{ij}$ for $(i,j) \in B$ by $CS_{ij} = \overrightarrow{CS}_{ij} + \overleftarrow{CS}_{ij}$.

\noindent\textbf{(4) Select additional lexicon by credit score.} The additional lexicon is selected in descending order of the $CS$ for each translation pair $(i,j)$. 

Based on the steps mentioned above, we append the \emph{annotated lexicon} with the \emph{additional lexicon} that contains high credit translation pairs.

This message passing mechanism is related to the bootstrap routine in~\citep{DBLP:conf/acl/AgirreLA18}. However, we select the credible translation pairs from the intersection, rather than union, of the forward and backward set of translation pairs. In this way, we guarantee the high quality of the additional lexicon.

\section{Semi-Supervision with Two-way Interaction}
In the previous section, we have presented two message passing mechanisms to enhance supervised BLI and OT-based unsupervised BLI by prior transformation $Q_{unsup}$ and $Q_{sup}$, respectively.
Moreover, recent state-of-the-art (SOTA) supervised($\supe$) and unsupervised($\unsup$) approaches are all based on stochastic optimization rather than the closed-form solution.
This means that all SOTA $\supe$ and $\unsup$ approaches can be considered as a module that operates on the feed-in parameter $Q$.
Therefore, we propose two different strategies for semi-supervision that emphasize the two-way interaction between the supervised signal and unsupervised alignment based on the message passing mechanisms, see Figure~\ref{fig:css-dss}.
All SOTA $\supe$ and OT-based $\unsup$ methods can be plugged into the proposed framework seamlessly.

\subsection{Cyclic Semi-Supervision}
The first proposed semi-supervised BLI strategy is \name, see Figure~\ref{fig:css-dss} (a).
\name feeds the parameter $Q$ into $\supe$ and $\unsup$ iteratively in a cyclic parameter feeding routine.
Cyclic parameter feeding is a ``hard'' way to share the parameters and is no more than~\citet{DBLP:conf/acl/PatraMGGN19} itself.
Besides parameter feeding, we propose to use the message passing mechanisms \BLU and \POT to strengthen the $\supe$ and $\unsup$.
However, there is no convergence guarantee for this optimization scheme.
As a result, it may suffer from limited performance when the BLI task is hard, as will be detailed in Section~\ref{sec:experiment}.

\subsection{Parallel Semi-Supervision}
The second strategy is \DSS, see Figure~\ref{fig:css-dss} (b), where $\supe$ and $\unsup$ are performed in parallel.
The information between $\supe$ and $\unsup$ is only passed by the proposed message passing mechanisms.
In this point of view, \citet{DBLP:conf/acl/ArtetxeLA17} only had the $\supe$ part with lexicon update and ignored the \unsup part.
Compared to \name, \DSS indirectly shares the information in a ``soft'' way and may be suitable for some hard BLI tasks. We use the metric formulated in Equation~\ref{eq:wasserstein-proc} to evaluate $Q_{sup}$ and $Q_{unsup}$ on the word embedding spaces and choose the better one as the final output of \DSS.

\begin{table*}[t]
	\small
	\begin{center}
		\renewcommand{\arraystretch}{1.2}
		\begin{tabular}{DCCCCCCCCCCCC}
			\toprule[2pt]
			\multirow{2}{3.5cm}{Method}           & \multicolumn{2}{c}{EN-ES} & \multicolumn{2}{c}{EN-FR} & \multicolumn{2}{c}{EN-DE} & \multicolumn{2}{c}{EN-RU} &  \multicolumn{2}{c}{EN-IT} & Avg.\\ \cline{2-13} &    $\rightarrow$    &    $\leftarrow$         &       $\rightarrow$      &      $\leftarrow$       &      $\rightarrow$       &      $\leftarrow$             &     $\rightarrow$        &    $\leftarrow$         &      $\rightarrow$       &      $\leftarrow$       \\\hline
			\textit{\textbf{Unsupervised Baselines}} & \multicolumn{11}{l}{}\\
			\citet{DBLP:conf/acl/AgirreLA18}    & 82.2        & 84.5        & 82.5        & \underline{83.6}        & 75.2        & 74.2       & \underline{48.5}        & \underline{65.1}    & 78.9  & \underline{79.5} & \underline{75.4}     \\	
			\citet{DBLP:conf/iclr/LampleCRDJ18}    & 81.7        & 83.3        & 82.3        & 82.1        & 74.0        & 72.2        & 44.0        & 59.1        &    78.3     & 78.1  &  73.5   \\				
			\citet{DBLP:conf/naacl/MohiuddinJ19} & 82.6        & 84.4        &       \underline{83.5}  &    82.4    & \underline{75.5}       & 73.9        &    41.2    & 61.7     & 78.8  & 78.5 & 74.2      \\
			\citet{DBLP:conf/aistats/GraveJB19}  & \underline{82.8}    & 84.1        & 82.6        & 82.9        & 75.4        & 73.3        & 43.7        & 59.1          &    66.6     & 62.5 &    71.3 \\
			\citet{DBLP:journals/corr/abs-1811-01124} & 82.4 & \underline{85.1}& 82.7 & 83.4 & \underline{75.5}& \underline{74.4} & 45.8 & 64.9 & \underline{79.4}& 79.4&  75.3 \\

			\hline
			
			\multicolumn{10}{l}{\textit{\textbf{Supervised Baselines with “5K all” annotated lexicon}}} & \multicolumn{2}{l}{}\\	
			\citet{DBLP:conf/emnlp/ArtetxeLA16} & 81.9 & 83.4 & 82.1 & 82.4 & 74.2 & 72.7 & 51.7 & 63.7 & 77.4 & 77.9 & 74.7 \\
			\citet{DBLP:conf/emnlp/JoulinBMJG18}& \underline{84.1} & \underline{86.3} & \underline{83.3} & \underline{84.1} & \textbf{\underline{79.1}} &76.3 &\textbf{\underline{57.9}} &67.2 & \underline{79.0}& \underline{81.4}& \underline{77.9}\\
			\citet{DBLP:journals/tacl/JawanpuriaBKM19}& 81.4 & 85.5 & 82.1 & \underline{84.1} & 74.7 & \underline{76.7} & 51.3 & \underline{67.6} & 77.8 & 80.9 & 76.2\\			
			 \hline
			
			\multicolumn{10}{l}{\textit{\textbf{Semi-Supervised Baselines with “100 unique” annotated lexicon}}} & \multicolumn{2}{l}{}\\
			\citet{DBLP:conf/acl/ArtetxeLA17}&79.9 & 83.2& 82.8 & 83.0&72.9&72.5& 38.9& 62.2& 78.5&77.7& 73.2\\
			\citet{DBLP:conf/acl/PatraMGGN19}&\textless 3 & \textless 3 & \textless 3 & \textless 3 & \textless 3 & \textless 3 & \textless 3 & \textless 3 & \textless 3 & \textless 3 & \textless 3 \\
			\name\textsf{- RCSLS} &\underline{83.9} & \underline{85.1} & \underline{83.7} & \underline{83.5} & \underline{77.5}  & \underline{74.6} & \underline{48.8} & \underline{63.0} &\underline{79.9} & \underline{80.5}  & \underline{76.0}\\
			\DSS\textsf{- RCSLS} &  82.0  & 83.1 & 82.1  & 81.9  & 74.4  & 72.2 & 46.5  & 61.5  & 78.7 &78.9 & 74.1\\
			\hline			
			\multicolumn{10}{l}{\textit{\textbf{Semi-Supervised Baselines with “5K unique” annotated lexicon}}} & \multicolumn{2}{l}{}\\
			\citet{DBLP:conf/acl/ArtetxeLA17}& 82.7& 83.3 & 82.9 & 83.3 & 75.9 & 72.4  & 47.6 & 62.3 & 78.7 & 77.7& 74.7 \\

			\citet{DBLP:conf/acl/PatraMGGN19}& 82.2 &84.6 & 82.6 & 83.9 & 75.6 & 73.7 &52.2 &65.2&77.8&78.6& 75.6\\
			\citet{DBLP:journals/corr/abs-2004-13889} & 80.9 & 80.8& - & - & 74.9 & 72.3&52.2 & 64.8 &77.1 &76.5& 72.4\\
			\name\textsf{- RCSLS} &\textbf{\underline{84.5}} &86.4 &\underline{84.5} &\underline{84.9} & \underline{78.8} & \underline{77.4} &\underline{57.0} &66.5 &\textbf{\underline{81.4}} &\underline{82.6}&\underline{78.4} \\
			
			\DSS\textsf{- RCSLS} & 83.5 & \underline{85.9} & 84.2 & 84.5 & 77.1 & 76.8 & 56.5 & \underline{67.1} & 80.0 & 82.1 & 77.8\\
			\hline
			\multicolumn{10}{l}{\textit{\textbf{Semi-Supervised Baselines with “5K all” annotated lexicon}}} & \multicolumn{2}{l}{}\\
			\citet{DBLP:conf/acl/ArtetxeLA17}& 82.3 & 83.5 & 82.9 & 82.7 & 76.3 & 72.5 & 48.7 & 62.3& 77.9 & 78.3& 74.7\\

			\citet{DBLP:conf/acl/PatraMGGN19}&84.3 & 86.2 & 83.9 & 84.7 & \textbf{\underline{79.1}} & 76.6 & 57.1 & 67.7 & 79.3 &82.4 & 78.1\\
				\citet{DBLP:journals/corr/abs-2004-13889} & 80.5 & 82.2 & - & - & 73.9 & 72.7 & 53.5 & 67.1 & 76.7 & 78.3 & 73.1\\
			\name\textsf{- RCSLS} & \textbf{\underline{84.5}}  & \textbf{\underline{86.9}}  & \textbf{\underline{85.3}}  & 85.3 &   78.9   & \textbf{\underline{78.7}}      &  \underline{57.3}       &  \textbf{\underline{67.9}}  &  \underline{81.2} & 82.7 &  \textbf{\underline{78.9}}   \\ 
			\DSS\textsf{- RCSLS} & 83.7& 86.5 & 84.4 & \textbf{\underline{85.5}} & 77.6 & 78.6 & 56.8 & 67.4 & 80.4 & \textbf{\underline{82.8}} & 78.4\\
			\bottomrule[2pt]     
			
		\end{tabular}
	\end{center}
	\caption{Word translation accuracy(@1) of \name and \DSS on the MUSE dataset with \textsf{RCSLS} as their supervised loss. ('EN': English, 'ES': Spanish, 'FR': French, 'DE': German, 'RU': Russian, 'IT': Italian. Underline: the highest accuracy among the group. In bold: the best among all methods).}
	\label{MUSE dataset}
	\vspace{-1em}
\end{table*}

\section{Experiment} \label{sec:experiment}
In this section, we conduct extensive experiments to evaluate the performance of \name and \DSS.
We open the source code on Github\footnote{\href{https://github.com/BestActionNow/SemiSupBLI}{https://github.com/BestActionNow/SemiSupBLI}}.

\subsection{Setup}
\noindent \textbf{Baselines}
We take several methods proposed in recent five years as baselines, including supervised~\citep{DBLP:conf/emnlp/ArtetxeLA16,DBLP:conf/emnlp/JoulinBMJG18,DBLP:journals/tacl/JawanpuriaBKM19}, unsupervised~\citep{DBLP:conf/emnlp/ArtetxeLA16,DBLP:conf/iclr/LampleCRDJ18,DBLP:conf/naacl/MohiuddinJ19,DBLP:conf/aistats/GraveJB19,DBLP:journals/corr/abs-1811-01124} and semi-supervised~\citep{DBLP:conf/acl/ArtetxeLA17,DBLP:conf/acl/PatraMGGN19,DBLP:journals/corr/abs-2004-13889} approaches. Brief introductions could be found in Section~\ref{sec:relatedwork}. 
The scores of baselines are retrieved from their papers or by running the publicly available codes if necessary. For \citet{DBLP:journals/corr/abs-2004-13889}, we don't find the released source code.

\noindent\textbf{Datasets}
We evaluate \name and \DSS against baselines on two popularly used datasets: the MUSE\footnote{\href{https://github.com/facebookresearch/MUSE}{https://github.com/facebookresearch/MUSE}} dataset \citep{DBLP:conf/iclr/LampleCRDJ18} and the VecMap\footnote{\href{https://github.com/artetxem/vecmap}{https://github.com/artetxem/vecmap}} dataset \citep{DBLP:journals/corr/DinuB14}. 
The MUSE dataset consists of FASTTEXT word embeddings \citep{DBLP:journals/tacl/BojanowskiGJM17} trained on Wikipedia corpora and more than 100 bilingual dictionaries of different languages. 
The FASTTEXT embeddings used in MUSE are trained on very large and highly semantically similar language
corpora (Wikipedia), which means the results on MUSE are biased~\citep{DBLP:conf/acl/AgirreLA18} and easier to obtain.
On the contrary, the VecMap dataset is less biased and harder using CBOW embeddings trained on the WacKy scrawled corpora and bilingual dictionaries obtained from the Europarl word alignments~\citep{DBLP:journals/corr/DinuB14}.
We use the default training and test splits for both datasets.

\noindent\textbf{Evaluation Setting} Similar to \citet{DBLP:journals/corr/abs-2004-13889}, we compare \name and \DSS against baselines on three annotated lexicons with different sizes, including one-to-one and one-to-many mappings: ``100 unique'' and “5K unique” contain one-to-one mappings of 100 and 5000 source-target pairs respectively, while “5K all” contains one-to-many mappings of all 5000 source and target words, that is, for each source word there may be multiple target words. Moreover, we present the experiment results of five totally unsupervised baselines and three supervised ones. All the accuracies reported in this section are the average of four repetitions. For detailed experimental data, such as the standard deviation, please refer to the tables in appendix.

\noindent\textbf{Hyperparameter Setting}
We train our models using Stochastic Gradient Descent with a batch size of 400 and a learning rate 1.0 for \supe, a batch size of 8K and a learning rate 500 for \unsup.
The temperature $T$ in Equation~\eqref{eq:prior-distribution} is 0.1 and the coefficient $\varepsilon$ in Equation~\eqref{eq:prior-ot} is 1. The additional lexicon size is set 10000. Each epoch contains 2K supervised iterations and 50 unsupervised iterations. Each case runs 5 epochs.
The aforementioned parameters work sufficiently good and we didn't search the best hyperparameters in this work.
All the experiments are conducted by 32-core CPU and one NVIDIA Tesla V100 core. 
Our framework finished in 30 minutes, while the running time for \citet{DBLP:conf/naacl/MohiuddinJ19} was 3 hours.

\subsection{Results on MUSE Dataset}
In Table~\ref{MUSE dataset}, we show the word translation results for five language pairs from the MUSE dataset, including 10 BLI tasks considering bidirectional translation.

With “100 unique” annotated lexicon, \name outperforms all other semi-supervised methods on every task. 
The accuracy score of \citet{DBLP:conf/acl/PatraMGGN19} is less than 3$\%$ on all tasks because the limited annotated lexicon is insufficient for effective learning, while \citet{DBLP:conf/acl/ArtetxeLA17} avoided this problem by lexicon bootstrap. 
Both \name and \DSS keep strong performance with insufficient annotated lexicon by the proposed message passing mechanisms, and achieve $2.8\%$ and $0.9\%$ improvement over \citet{DBLP:conf/acl/ArtetxeLA17}, respectively.
Compared to the iterative \name that feeds parameters by \unsup directly into \supe, the parallel \DSS has fewer connections between \supe and \unsup. Thus, \name performance is better than \DSS under low supervision.

We notice that semi-supervised approaches with "100 unique" annotated lexicon are even worse than the unsupervised methods. This indicates that 100 annotation lexicon is too weak for supervised approach to learn meaningful transformation.  It does not mean our approach has marginal contribution. On the contrary, these empirical results reveal that bad supervised BLI won't hurt the overall performance of our semi-supervised framework and this is what previous work cannot achieve.

As the annotated lexicon size increases, the dominance of \name and \DSS is still observed. Moreover, the gap between \name and \DSS disappears as the size of annotated lexicon gets larger. 
With ``5K unique'' annotated lexicon, \name and \DSS outperform other semi-supervised methods on all tasks. 
With ``5K all'' annotated lexicon, \name and \DSS outperform other semi-supervised baselines on 9 of 10 tasks. 
On average, \name exceeds  \citet{DBLP:conf/acl/ArtetxeLA17}, \citet{DBLP:conf/acl/PatraMGGN19}
and \citet{DBLP:journals/corr/abs-2004-13889} by $4.2\%, 0.8\%, 5.8\%$, respectively.

Taking all methods into consideration, including supervised, semi-supervised and unsupervised, \name and \DSS achieve the highest accuracy on 8 of 10 tasks and the best results on average.

\subsection{Results on VecMap Dataset}
\begin{table}[t]
	\scriptsize
	\tiny
	\begin{center}
		\renewcommand{\arraystretch}{1.2}
		\setlength{\arraycolsep}{1pt}
		\begin{tabular}{EBBBBBBB}
			\toprule[2pt]
			\multirow{2}{2.8cm}{Method}    & \multicolumn{2}{c}{EN-ES}  & \multicolumn{2}{c}{EN-IT} &   \multicolumn{2}{c}{EN-DE} & Avg.\\\cline{2-8}
			&    $\rightarrow$    &    $\leftarrow$         &       $\rightarrow$      &      $\leftarrow$       &      $\rightarrow$       &      $\leftarrow$             \\ \hline
			\textit{\textbf{Unsupervised Baselines}} & \multicolumn{7}{l}{}\\
			\citet{DBLP:conf/acl/AgirreLA18}    & 36.9 & 31.6& 47.9 &42.3 &\underline{48.3} &\underline{44.1} & \underline{41.9}\\		
			\citet{DBLP:conf/iclr/LampleCRDJ18}   & 34.7 & 0.0 &44.9 &38.7 &0.0 &0.0 & 19.7\\	
			\citet{DBLP:conf/naacl/MohiuddinJ19}& \underline{37.4} &31.9 &47.6 &42.5 &0.0 &0.0& 26.6\\
			\citet{DBLP:conf/aistats/GraveJB19} & 0.0 & 0.7 & 40.3 & 34.8& 0.0 & 37.1 & 18.8\\
			\citet{DBLP:journals/corr/abs-1811-01124}& 0.0& \textbf{\underline{58.3}} & \textbf{\underline{70.0}} & \textbf{\underline{69.5}} & 0.0 & 0.0 & 33.0\\
            \hline
			
			\multicolumn{6}{l}{\textit{\textbf{Supervised Baselines with “5K all” annotated lexicon}}} & \multicolumn{2}{l}{}\\	
			\citet{DBLP:conf/emnlp/ArtetxeLA16} & 19.5 & 13.7 & 39.3 & 20.7 & 25.4 & 22.3 & 23.5 \\
			\citet{DBLP:conf/emnlp/JoulinBMJG18}&35.5& 31.2 &44.6& 37.6& 46.6 &41.7& 39.5\\
			\citet{DBLP:journals/tacl/JawanpuriaBKM19}&\underline{37.5} &\underline{33.1} &\underline{47.6} &\underline{40.1} &\underline{48.8} & \textbf{\underline{45.1}} & \underline{42.0} \\
			\hline
			\multicolumn{6}{l}{\textit{\textbf{Semi-Supervised Baselines with “100 unique” annotated lexicon}}} & \multicolumn{2}{l}{}\\
			\citet{DBLP:conf/acl/ArtetxeLA17}&33.1 & 24.9 & 43.3 & 39.2 & 46.9 & 42.0 & 38.2 \\
			\citet{DBLP:conf/acl/PatraMGGN19}& \textless 3 & \textless 3 & \textless 3 & \textless 3 & \textless 3 & \textless 3 & \textless 3\\
			\name\textsf{- RCSLS}  
			&\underline{36.8} &\underline{31.4} &\underline{45.4} &\underline{40.9} &\underline{48.0} &42.2 & \underline{40.8}\\ 
			\DSS\textsf{- RCSLS} &34.6 &29.6 &45.3 &40.5 &\underline{48.0} &\underline{42.6}&40.1\\
			\hline			
			\multicolumn{6}{l}{\textit{\textbf{Semi-Supervised Baselines with “5K unique” annotated lexicon}}} & \multicolumn{2}{l}{}\\
			\citet{DBLP:conf/acl/ArtetxeLA17}&33.3 &27.6 &43.9 &38.4 &46.0 &41.1 & 38.4\\
			\citet{DBLP:conf/acl/PatraMGGN19}&34.3& 31.6 &41.1 &39.3 &47.5 &43.6&39.6\\
				\citet{DBLP:journals/corr/abs-2004-13889} &33.4& 27.3 &44.1 &38.9 &42.5 &39.4 & 37.6\\
			\name\textsf{- RCSLS} &38.1 &32.2 &46.4 &\underline{41.2} &47.9 &43.2 & 41.5\\ 
			\DSS\textsf{- RCSLS} &\textbf{\underline{38.9}}$^\dagger$ & \underline{32.9} & \textbf{\underline{47.8}}$^\dagger$ & 41.1 & \textbf{\underline{49.3}}$^\dagger$ & \underline{43.7} & \textbf{\underline{42.3}}$^\dagger$\\
			\hline
			\multicolumn{6}{l}{\textit{\textbf{Semi-Supervised Baselines with “5K all” annotated lexicon}}} & \multicolumn{2}{l}{}\\
			\citet{DBLP:conf/acl/ArtetxeLA17}&32.7 & 28.1 & 43.8 & 38.0 &47.4 &40.8& 38.5\\
			\citet{DBLP:conf/acl/PatraMGGN19}&34.5 &32.1 &46.2 &39.5 &48.1 &44.1& 40.8\\
				\citet{DBLP:journals/corr/abs-2004-13889} &33.7 &27.9 &43.7 &38.9 &43.6 & 39.2 & 37.8\\
			\name\textsf{- RCSLS} &38.9 &32.5 &46.6 &41.3 &48.4 &42.5 & 41.7\\ 
			\DSS\textsf{- RCSLS} & \textbf{\underline{39.6}} & \textbf{\underline{33.7}}$^\dagger$ & \textbf{\underline{47.8}}$^\dagger$ & \textbf{\underline{42.1}}$^\dagger$ & \textbf{\underline{50.8}} & \textbf{\underline{44.8}}$^\dagger$& \textbf{\underline{43.1}}\\ 
			\bottomrule[2pt]     
			
		\end{tabular}
	\end{center}
	\caption{Word translation accuracy(@1) of \name and \DSS on the VecMap dataset with \textsf{RCSLS} as their supervised loss. ('EN': English, 'ES': Spanish,  'DE': German, 'IT': Italian. Underline: the highest accuracy among the group. In bold: the best among all methods. In bold and marked by $\dagger$: the second-highest among all methods).}
    \label{VecMap dataset}
    \vspace{-5em}
\end{table}

\begin{table*}[]
	\tiny
	\begin{center}
		\renewcommand{\arraystretch}{1.2}
		\begin{tabular}{X|YYYY|YYYY|Y||YYYY|YYYY|Y||Y}
			\toprule[2pt]
            \hline
            \multicolumn{1}{l|}{Annotated Lexicon Size} & \multicolumn{9}{c||}{5K all} & \multicolumn{9}{c||}{1K unique} & \multicolumn{1}{c}{\multirow{4}{0.3cm}{\tabincell{c}{avg\\of all}}}\\
            \cline{1-19}
		    \multicolumn{1}{l|}{Dataset}&\multicolumn{4}{c|}{MUSE} &\multicolumn{4}{c|}{VecMap}&\multicolumn{1}{c||}{\multirow{3}{0.3cm}{avg}}
		    &\multicolumn{4}{c|}{MUSE} &\multicolumn{4}{c|}{VecMap}&\multicolumn{1}{c||}{\multirow{3}{0.3cm}{avg}} &\\
			\cline{1-9} \cline{11-18}
			\multirow{2}{2cm}{Languages}  & \multicolumn{2}{c}{EN-ES} & \multicolumn{2}{c|}{EN-FR} & \multicolumn{2}{c}{EN-DE} & \multicolumn{2}{c|}{EN-IT}&& \multicolumn{2}{c}{EN-ES} & \multicolumn{2}{c|}{EN-FR} & \multicolumn{2}{c}{EN-DE} & \multicolumn{2}{c|}{EN-IT}&\\ 
			\cline{2-9} \cline{11-18}  
			 &$\rightarrow$&$\leftarrow$&$\rightarrow$&$\leftarrow$&$\rightarrow$&$\leftarrow$&$\rightarrow$&$\leftarrow$&&$\rightarrow$&$\leftarrow$&$\rightarrow$&$\leftarrow$&$\rightarrow$&$\leftarrow$&$\rightarrow$&$\leftarrow$&\\ \hline
			\multicolumn{10}{l||}{\textit{\textbf{Results of the Ablation to \name}}} & \multicolumn{9}{l||}{\textit{\textbf{Results of the Ablation to \name}}}\\
            \arrayrulecolor{black} \cdashline{1-20}[1.5pt/2pt]
			\name\textsf{- RCSLS}   &84.5&86.9&85.5&85.3& 48.4 & 42.5& 46.6 & 41.3  & 65.1&83.8 &85.0 &83.9 &83.7 &47.8 &42.8 &45.3 &41.2 & 64.2 & 64.7\\
    		
			$\circleddash$ \POT  &83.9 &86.6 &84.0 &84.8&46.6 &41.7 &43.7 &39.5 & 63.9 &81.6 &84.8 &82.2 &83.4 &41.8 &40.2 &36.5 &35.7 &60.8  & 62.3\\

			$\circleddash$ \BLU   &83.2 &86.6 &84.4 &84.7& 47.4 &42.9 &45.4 &40.4 & 64.4&82.5 &83.7 &82.5 &82.2 &47.7 &42.6 &45.3 &39.7 & 63.3 & 63.8\\

			$\circleddash$ \POT $\&$ $\circleddash$ \BLU
    			   &82.5 &84.9 &83.0 &83.7 &41.7 &36.6 &39.7 &34.8 & 60.9 &61.0 &62.4 &57.4 &59.5 &28.1 &22.4 &26.2 &23.7 & 42.6 & 51.7\\
            \arrayrulecolor{black} \cdashline{1-20}[1.5pt/2pt]
			$\circleddash$ \unsup $\&$ $\circleddash$ \POT 
    			   &84.3 &86.5 &84.8 &85.1 &45.8 &40.1 &42.8 &39.0 & 63.6& 81.7 &83.3 &80.5 &81.4 &40.3 &36.1 &38.2 &36.4 & 59.7 & 61.6\\
 
			$\circleddash$ \supe $\&$ $\circleddash$ \BLU 
    			  &82.3 &83.2 &82.5 &82.7& 47.7 &43.2 &45.3 &40.4 & 63.4 &82.5 &83.8 &82.2 &82.9 &47.8 &42.8 &45.2 &39.7 & 63.4 & 63.4\\
			\hline	
			\multicolumn{10}{l||}{\textit{\textbf{Results of the Ablation to \DSS}}} & \multicolumn{9}{l||}{\textit{\textbf{Results of the Ablation to \DSS}}}\\
			\arrayrulecolor{black} \cdashline{1-20}[1.5pt/2pt]
			\DSS\textsf{- RCSLS}   &83.7 & 86.5 & 84.4 & 85.5 & 50.8 & 44.8 & 47.8 & 42.1 & 65.7 &82.9 &83.8 &82.4 &83.0 &48.4 &43.0 &46.6 &40.1 & 63.8 & 64.7\\
			$\circleddash$ \POT   &83.5 &85.4 &84.4 &85.3 &49.1 &43.1 &46.5 &40.8 & 64.8 &81.1 &82.6 &82.4 &81.9 &45.1 &40.7 &41.7 &36.4 & 61.5 & 63.1\\
			$\circleddash$ \BLU   &82.8 &85.4 &83.0 &84.3 &48.5 &43.7 &46.0 &39.9 & 64.2 &81.9 &83.9 &82.2 &82.5 &48.0 &42.6 &44.8 &39.0 & 63.1 & 63.7\\
    		\hline
			\bottomrule[2pt]     
		\end{tabular}
	\end{center}
	 \caption{Ablation Study with "5K all" and "1K unique" annotated lexicon. ($\circleddash$: remove specific component from the \name or \DSS. $\&$: remove both components.)}
	\label{Ablation}
	\vspace{-1em}
\end{table*}

In Table~\ref{VecMap dataset}, we show the word translation accuracy for three language pairs, including 6 translation tasks on the harder VecMap dataset~\citep{DBLP:journals/corr/DinuB14}.

Notably, a couple of unsupervised approaches ~\citep{DBLP:conf/iclr/LampleCRDJ18,DBLP:conf/naacl/MohiuddinJ19,DBLP:conf/aistats/GraveJB19,DBLP:journals/corr/abs-1811-01124} are evaluated to have a zero accuracy on some of the language pairs.
On the one hand, their valotile results demonstrate the toughness of the VecMap dataset where the structural similarity for unsupervised BLI is very low.
On the other hand, unstable performance may be explained by the high dependence of those methods on the initialization.
Though the performance of those methods are highest in some cases, e.g.~\citet{DBLP:journals/corr/abs-1811-01124}, due to the unstable nature. We also mark the second-highest score by bold font and $\dagger$ if necessary.

At all supervision levels, \name and \DSS outperform all other semi-supervised approaches.
Taking all unsupervised, semi-supervised and supervised methods into account, \name and \DSS achieve SOTA accuracy on average. 
Notably, \DSS gets the highest or the second-highest (except the unstable unsupervised baseline~\citep{DBLP:journals/corr/abs-1811-01124}) scores for 5 of 6 language pairs.

\vspace{-0.5em}
The results for “100 unique” annotated lexicon support our finding on the MUSE dataset that \name learns better at low supervision level.
Interestingly, with “5K unique” and “5K all” annotated lexicons, \DSS outperforms \name on almost every task, which is different from the MUSE dataset. Given that the structural similarity of embeddings between different languages in VecMap is very low, \unsup procedure is very unstable.
In this case, \name has lower performance due to the unstable $Q_{unsup}$ is directly fed into \supe, while the parallel strategy of \DSS does not suffer from this problem. 

\subsection{Ablation Study}
\label{Sec5.4}

In the ablation study, we disassemble \name and \DSS into the basic components to analyze the contribution of each component. 
Specifically, we consider the proposed two message passing mechanisms \POT and \BLU. For \name, we also include the effect of \supe or \unsup module in the cyclic parameter feeding. However, if \supe or \unsup in \DSS is removed, the framework falls back to the unsupervised or supervised BLI, whose results are already in Table~\ref{MUSE dataset} and~\ref{VecMap dataset}.

We conduct ablation experiments with two annotated lexicons with different sizes, "5K all" and "1K unique" to compare the behavior of \name and \DSS under different annotation level.
The experimental setting is the same as the main experiments. 
The ablation results are presented in Table~\ref{Ablation} on four language pairs (2 from MUSE dataset and 2 from VecMap dataset). 

\noindent \textbf{Effectiveness of \POT and \BLU:}

Regardless of the annotated lexicon size, removing \POT, \BLU and both of them from \name brings $2.4\%$, $0.9\%$ and $13.0\%$ decline of accuracy respectively on average. Notably, the cyclic parameter feeding does not bring further benefits.
Only when combined with at least one message passing mechanism, \POT or \BLU, the accuracy is improved significantly. 
For \DSS, removal of \POT or \BLU brings $1.6\%$ and $1.0\%$ decline on the average score respectively.


Moreover, we consider different annotated lexicon sizes. 
On average, removal of \POT, \BLU or both from \name brings $1.2\%$, $0.7\%$ and $4.2\%$ decline respectively with "5K all" annotated lexicon size, $3.4\%$, $0.9\%$ and $21.6\%$ decline with "1K unique" annotated lexicon size. 
The message passing mechanisms contribute drastically with a smaller annotated lexicon size since \supe receives significantly larger additional lexicons from \unsup to strengthen its performance. 
As for \DSS, removal of \POT and \BLU brings $0.9\%$ and $1.5\%$ decline respectively with "5K all" annotated lexicon size, $2.3\%$ and $0.7\%$ decline with "1K unique" annotated lexicon size. 
No significant effect of annotation level for \DSS is observed in ablation study.
For both \name and \DSS, the contribution of \POT is slightly larger than that of \BLU and the combination of them could bring impressive improvement in general.

\noindent \textbf{Analysis of \supe and \unsup in \name:}

In this step, we remove \supe or \unsup from \name and monitor the performance change. 
Note that if we remove \unsup from \name, \POT also needs to be removed as we do not need any prior transport plan for \unsup anymore. 
Removing \supe also means the removal of \BLU for a similar reason.
After removing \unsup and \POT, \name feeds $Q_{sup}$ exactly to \BLU for additional lexicon and then to \supe again, just like \citet{DBLP:conf/acl/ArtetxeLA17, DBLP:conf/acl/AgirreLA18}. After removing \supe and \BLU, \unsup takes the transformation learned by itself in previous steps to generate the prior transport plan. 
The average accuracy drops by $1.5\%$ and $4.5\%$ with "5K all" and "1K unique" annotated lexicon respectively after removing \unsup, by $1.7\%$ and $0.8\%$ after removing \supe.


Given the comparison above, \supe contributes less than \unsup with "1K unique" annotated lexicon. 
Whereas \supe and \unsup contribute comparably with "5K all" annotated lexicon.
In other words, at low annotation level, i.e. "1K unique", where \supe BLI does not work well, the participation of \unsup extends the valuable additional lexicon.

\subsection{Results on distant language pairs}

\begin{table*}[t]
	\small
	\begin{center}
		\renewcommand{\arraystretch}{1.2}
		\begin{tabular}{DCCCCCCCCCCCC}
			\toprule[2pt]
			\multirow{2}{3.5cm}{Method}           & \multicolumn{2}{c}{EN-ZH} & \multicolumn{2}{c}{EN-TA} & \multicolumn{2}{c}{EN-JA} & \multicolumn{2}{c}{EN-MS} &  \multicolumn{2}{c}{EN-FI} & Avg.\\ \cline{2-13} &    $\rightarrow$    &    $\leftarrow$         &       $\rightarrow$      &      $\leftarrow$       &      $\rightarrow$       &      $\leftarrow$             &     $\rightarrow$        &    $\leftarrow$         &      $\rightarrow$       &      $\leftarrow$       \\\hline
			
            \multicolumn{10}{l}{\textit{\textbf{(Semi-)Supervised Baselines with “5K unique” annotated lexicon}}} & \multicolumn{2}{l}{}\\
            
			\citet{DBLP:conf/acl/PatraMGGN19}    &   42.5    &   42.8      &    15.3     &   22.0      &   3.3     &   34.8     &  50      &  49.3  & 48.6  &  60.9  & 37.0  \\
			
			\citet{DBLP:journals/tacl/JawanpuriaBKM19}    &   43.7      &    40.1     &       16.1  &    22.0     &  0.0      &        0.0 &     51.0   &    49.8 & 47.4  & 65.1 & 33.5      \\
			
			\citet{DBLP:conf/naacl/ZhouMWN19}    &   42.5      &   40.0      &     \textbf{17.9}    &   \textbf{27.9}      &     \textbf{52.0}   &    35.6    &   37.6     & 40.1   & 50.7  & 60.8 & 40.5      \\

			\name\textsf{- RCSLS}  &\textbf{48.6} &\textbf{46.0} &17.7 &24.5 &50.3 &\textbf{41.9} &\textbf{54.6} &55.5 &\textbf{56.0} &65.6  & \textbf{46.1}\\	
			
			\DSS\textsf{- RCSLS} &47.8 &45.4 &17.3 &23.3 &50.6 &41.3 &54.6 &\textbf{55.6} &55.6 &\textbf{66.4}   &45.8   \\	
			\bottomrule[2pt]     
			
		\end{tabular}
	\end{center}
	\caption{Word translation accuracy(@1) of \name and \DSS on the distant language pairs with \textsf{RCSLS} as their supervised loss. ('EN': English, 'TA': Tamil, 'JA': Japanese, 'MS': Malay, 'FI': Finnish. In bold: the best among all methods).}
	\label{dis-L}
	\vspace{-0.4cm}
\end{table*}

In this section, We report the tranlation accuracy of our method on five distant language pairs with 5000 lexicon. We choose three methods as baselines: \citet{DBLP:conf/acl/PatraMGGN19} proposed semi-supervised SOTA method. \citet{DBLP:journals/tacl/JawanpuriaBKM19} is the supervised SOTA method. \citet{DBLP:conf/naacl/ZhouMWN19} designed an unsupervised  matching procedure with density matching technologies, which achieved significant improvement on distant language pairs.  As we need to compare supervised, unsupervised and semi-supervised method  simultaneously, we conduct evaluation only on the "5K unique" supervision level. 

As shown in Table~\ref{dis-L}, our method also retains a distinct advantage on these distant language pairs. In the cases between "EN" and "JA", \citet{DBLP:conf/acl/PatraMGGN19} and \citet{DBLP:journals/tacl/JawanpuriaBKM19} are completely inefficient. While our method obtains stable results on these cases, which proves the robustness of \name and \DSS. Moreover, our method outperforms \citet{DBLP:conf/naacl/ZhouMWN19} on most cases. In short, \name and \DSS could obtain stable and better results on various language pairs.


\section{Related Work}\label{sec:relatedwork}
This paper is mainly related to the following three lines of work.

\noindent\textbf{Supervised methods.} \citet{DBLP:journals/corr/MikolovLS13} pointed out that it was a feasible way to BLI by learning a linear transformation based on the Euclidean distance. \citet{DBLP:conf/emnlp/ArtetxeLA16} applied normalization to word embeddings and imposed an orthogonal constraint on the linear transformation which led to a closed-form solution. \citet{DBLP:conf/emnlp/JoulinBMJG18} replaced Euclidean distance with the RCSLS distance to relieve the hubness phenomenon and achieved SOTA results for many languages. \citet{DBLP:journals/tacl/JawanpuriaBKM19} optimized a Mahalanobis metric along with the transformation to refine the similarity between word embeddings.

\noindent\textbf{Unsupervised methods.} 
\citet{DBLP:conf/acl/AgirreLA18} proposed an unsupervised method to generate an initial lexicon by exploiting the similarity in cross-lingual space and applied a robust self-learning to improve it iteratively.
~\citet{DBLP:conf/iclr/LampleCRDJ18} did the first work for unsupervised BLI which learned a linear transformation by adversarial training and improved it by a refinement procedure.
~\citet{DBLP:conf/naacl/MohiuddinJ19} revisited adversarial autoencoder for unsupervised word translation and proposed two novel extensions to it.
Moreover, OT-based unsupervised BLI is the central part in this paper.
~\citet{DBLP:conf/emnlp/Alvarez-MelisJ18} exploited the structure similarity of embedding space by minimizing the Gromov-Wasserstein metric between source and target word embedding distributions.
~\citet{DBLP:conf/aistats/GraveJB19} viewed unsupervised BLI task as the minimization of Wasserstein distance between the source and target distributions of word embeddings. They optimized this problem by using Sinkhorn and Procrustes alternatively.
~\citet{DBLP:journals/corr/abs-1811-01124} furthered the work of ~\citet{DBLP:conf/aistats/GraveJB19} by using the RCSLS as the distance metric, which addresses hubness phenomenon better than Euclidean distance.
~\citet{xuzhao2020} proposed an relaxed matching procedure derived from unbalanced OT algorithms and solved the polysemy problem to a certain extent.
~\citet{DBLP:conf/emnlp/XuYOW18} used a neural network implementation to calculate the Sinkhorn distance, a well-defined OT-based distributional similarity measure, and optimized the objective through back-propagation. 

\noindent\textbf{Semi-supervised methods.} 
\citet{DBLP:conf/acl/ArtetxeLA17} proposed a simple self-learning
approach that can be combined with any dictionary-based mapping technique and started with almost no lexicon. \citet{DBLP:conf/acl/PatraMGGN19} proposed a semi-supervised approach that relaxes the isometric assumption and optimizes a supervised loss and an unsupervised loss together.

Notably, comparing with the self-learning method like \citep{DBLP:conf/acl/AgirreLA18} or \citep{DBLP:conf/emnlp/VulicGRK19}, our framework with two message passing mechanisms is quite different from theirs. Although the lexicon updating procedures in their papers are similar with the \BLU that we proposed, there are two main differences:
(1) Their approaches use the lexicon from current step to extract the lexicon for next step. Meanwhile, BLU uses unsupervised output to extract lexicon for the supervised part. Our models will degenerate to their situation after removing the unsupervised part and POT. This situation has been discussed in the ablation study in Section~\ref{Sec5.4}.
(2) BLU extracts lexicon according to bidirectional matching information while they only consider one direction. This trick improves the lexicon quality.

Moreover, alignment of word embeddings in latent spaces by Auto-Encoders or other projections is another trend of BLI research.
Latent space alignment includes unsupervised variants~\citep{DBLP:conf/emnlp/DouZH18,
DBLP:journals/taslp/BaiCCZ19, DBLP:conf/naacl/MohiuddinJ19} and semi-supervised variants~\citep{DBLP:journals/corr/abs-2004-13889}. 
We emphasize that the latent space alignment is orthogonal to our proposed framework.
Our entire framework can be transferred to any given latent space. 

\section{Conclusions}
In this paper, we introduce the two-way interaction between the supervised signal and unsupervised alignment by proposed \POT and \BLU message passing mechanisms. \POT guides the OT-based unsupervised BLI by prior BLI transformation. \BLU employs a bidirectional retrieval to enlarge the annotated data and stabilize the training of supervised BLI approaches. Ablation study shows that the two-way interaction by \POT and \BLU is the key to significant improvement. 

Based on the message passing mechanisms, we design two strategies of semi-supervised BLI to integrate supervised and unsupervised approaches, \name and \DSS, which are constructed on cyclic and parallel strategies respectively. The results show that \name and \DSS achieve SOTA results over two popular datasets. 
As \name and \DSS are compatible with any supervised BLI and OT-based unsupervised BLI approaches, they can also be applied to the latent space optimization.

\bibliography{emnlp2020}
\bibliographystyle{acl_natbib}

\clearpage
\newpage
\onecolumn
\section*{Appendix}
\setcounter{table}{0}

\begin{center}
\begin{table*}[ht]
  \tiny
  \centering
  \renewcommand{\arraystretch}{1.2}
  
  \resizebox{\textwidth}{!}{
		\begin{tabular}{XY|YYYYYYYYYY||YYYYYY||YYYYYYYYYY}
			\hline
			\multicolumn{2}{l|}{Dataset}  & \multicolumn{10}{c||}{MUSE} & \multicolumn{6}{c||}{VecMap} &
			\multicolumn{10}{c}{Distant Language Pairs}\\
			\hline
			\multicolumn{2}{l|}{\multirow{2}{2cm}{Languages}}    & \multicolumn{2}{c}{EN-ES} & \multicolumn{2}{c}{EN-FR} & \multicolumn{2}{c}{EN-DE} & \multicolumn{2}{c}{EN-RU} &  \multicolumn{2}{c||}{EN-IT} & \multicolumn{2}{c}{EN-ES} & \multicolumn{2}{c}{EN-IT} &  \multicolumn{2}{c||}{EN-DE} &
			\multicolumn{2}{c}{EN-ZH} & \multicolumn{2}{c}{EN-TA} & \multicolumn{2}{c}{EN-JA} & \multicolumn{2}{c}{EN-MS} &  \multicolumn{2}{c}{EN-FI}\\ 
			\cline{3-28}     
			& &$\rightarrow$&$\leftarrow$&$\rightarrow$&$\leftarrow$&$\rightarrow$&$\leftarrow$&$\rightarrow$&$\leftarrow$&$\rightarrow$&$\leftarrow$&$\rightarrow$&$\leftarrow$&$\rightarrow$&$\leftarrow$&$\rightarrow$&$\leftarrow$\\ \hline
			\multicolumn{12}{l||}{\textit{\textbf{Semi-Supervised Baselines with “100 unique” word dictionary}}} & \multicolumn{6}{l||}{}\\
			\arrayrulecolor{black} \cdashline{1-28}[1.5pt/2pt]
			\multirow{3}{2cm}{\name\textsf{- RCSLS}} & \textbf{best} & 84.1 & 85.2 & 83.9 & 83.5 & 77.7 & 74.7 & 49.3 & 63.1 & 80.1 & 80.7 &36.9 &32.8 &45.9 &41.1 &48.5 &42.3\\
    			 & \textbf{avg} &83.9 &85.1 & 83.7  &83.5 &77.5 &74.6 &48.8 &63.0 &79.9 &80.5&36.8 &31.4 &45.4 &40.9 &48.0 &42.2 \\
    			 & \textbf{st} &0.193 &0.158 & 0.240 &0.065 &0.285 &0.122 &0.525 &0.138 &0.175 &0.140&0.205 &0.930 &0.341 &0.180 &0.375 &0.087 \\
    			 \arrayrulecolor{black} \cdashline{1-28}[1.5pt/2pt]
			\multirow{3}{2cm}{\DSS\textsf{- RCSLS}} & \textbf{best}& 82.2 & 83.3 & 82.6 & 82.1 & 74.87 & 72.6 & 47.2 & 61.6 & 78.7 & 79 & 35.0 &29.7 &45.7 &40.7 &48.3 &42.7 \\
    			 & \textbf{avg} & 82.0  & 83.1 & 82.1  & 81.9  & 74.4  & 72.2 & 46.5  & 61.5  & 78.7 &78.9 &34.6 &29.6 &45.3 &40.5 &48.0 &42.6\\
    			 & \textbf{st} &0.138 & 0.167 & 0.366 & 0.249 & 0.358 & 0.345 & 0.627 & 0.136 &0.078 &0.115 &0.553 &0.156 &0.406 &0.178 &0.381 &0.181\\
			\hline	
			\multicolumn{12}{l||}{\textit{\textbf{Semi-Supervised Baselines with “5K unique” word dictionary}}} & \multicolumn{6}{l||}{}\\
			\arrayrulecolor{black} \cdashline{1-28}[1.5pt/2pt]
			\multirow{3}{2cm}{\name\textsf{- RCSLS}} & \textbf{best} &84.7 &86.5 & 84.6 & 85.1 & 79.0 & 77.7 & 57.5 & 66.8 & 81.6 & 82.7 &38.4 &32.6 &46.5 &41.4 &48.3 &43.3 &48.7 &46.1 &18.1 &24.9 &50.5 &42.7 &55.1 &56.3 &56.5 &65.7\\
    			 & \textbf{avg} &84.5 &86.4 &84.5 &84.9 &78.8 &77.4 &57.0 &66.5 &81.4 &82.6&38.1 &32.2 &46.4 &41.2 &47.9 &43.2 &48.6 &46.0 &17.7 &24.5 &50.3 &41.9 &54.6 &55.5 &56.0 &65.6\\
    			 & \textbf{st} &0.167 &0.050 &0.115 &0.201 &0.158 &0.236 &0.817 &0.339 &0.168 &0.136 &0.318 &0.358 &0.177 & 0.313 & 0.369 & 0.113 &0.196 &0.103 &0.335 &0.524 &0.149 &0.729 &0.409 &0.717 &0.578 &0.191\\
    			 \arrayrulecolor{black} \cdashline{1-28}[1.5pt/2pt]
			\multirow{3}{2cm}{\DSS\textsf{- RCSLS}} & \textbf{best}& 83.7 & 86.0 & 84.5 & 84.7 & 77.2 & 77.0 & 56.7 & 67.6 & 80.1 & 82.1&
			39.1  & 33.2  & 48.2 & 41.3 & 49.6 & 43.8 &47.9 &45.4 &17.3 &23.4 &50.8 &41.6 &54.9  &55.8 &55.9 &66.6\\
    			 & \textbf{avg} & 83.5 & 85.9 & 84.2 & 84.5 & 77.05 & 76.8 & 56.5 & 67.1 & 80.0 & 82.1 &38.9 & 32.9 & 47.8 & 41.1 & 49.3 & 43.7 &47.8 &45.4 &17.3 &23.3 &50.6 &41.3 &54.6 &55.6 &55.6 &66.4
    			 \\
    			 & \textbf{st} & 0.173 & 0.115 & 0.236 & 0.171 & 0.129 & 0.183 & 0.283 & 0.395 & 0.082 & 0.100 &0.168 & 0.303 & 0.364 & 0.308 & 0.214 & 0.100 &0.075 &0.075 &0.115 &0.107 &0.171 &0.283 &0.196 &0.231 &0.273 &0.236 
    			 \\
			\hline
			\multicolumn{12}{l||}{\textit{\textbf{Semi-Supervised Baselines with “5K all”  word dictionary}}} & \multicolumn{6}{l||}{}\\
			\arrayrulecolor{black} \cdashline{1-28}[1.5pt/2pt]
			\multirow{3}{2cm}{\name\textsf{- RCSLS}} & \textbf{best} & 84.6 &  87.1 & 85.5& 85.3 & 78.9 & 78.8 & 57.8 & 68.2 & 81.4 & 82.8 &39.0 &32.7 &47.0 &41.4 &49.1 &42.9 \\
    			 & \textbf{avg} & 84.5  & 86.9 & 85.3 & 85.3 & 78.9 & 78.7 & 57.3 & 67.9 & 81.2 & 82.7 &38.9 &32.5 &46.6 &41.3 &48.4 &42.5\\
    			 & \textbf{st} & 0.126 & 0.481 & 0.238 & 0.099 & 0.087 & 0.115 & 0.521 & 0.285 & 0.198 & 0.053 &0.175 &0.295 &0.425 &0.178 &0.500 &0.272\\ 
    			 \arrayrulecolor{black} \cdashline{1-28}[1.5pt/2pt]
			\multirow{3}{2cm}{\DSS\textsf{- RCSLS}} & \textbf{best} &83.9 & 87.0 & 84.6 & 85.5 & 77.7 & 78.8 & 57.2 & 67.5 & 80.7 & 82.9 & 39.7 & 34.0 & 48.3 & 42.5 & 51.2 & 45.1 \\
    			 & \textbf{avg} & 83.7 & 86.5 & 84.4 & 85.5 & 77.6 & 78.6 & 56.8 & 67.4 & 80.4 & 82.8 &39.6 & 33.7 & 47.8 & 42.1 & 50.8 & 44.8\\
    			 & \textbf{st} & 0.163 & 0.359 & 0.150 & 0.100 & 0.115 & 0.206 &0.271 & 0.191 & 0.25 & 0.115 &0.196 & 0.368 & 0.968 & 0.436 & 0.513 & 0.345\\
        
			\hline
  \end{tabular}}

\caption{Detailed Experimental Results of \name and \DSS on MUSE Dataset, VecMap Dataset and distant language pairs. We repeat the experiment on each language pair four times and report \textbf{best}, \textbf{avg}, \textbf{st} of the four results(\textbf{best}: the highest @1 accuracy. \textbf{avg}: the average accuracy which is reported in main body of this paper. \textbf{st}: the standard deviation.)}
  \label{AT1}
\end{table*}
\end{center}

 \begin{table*}[h]
	\tiny
	\centering
		\renewcommand{\arraystretch}{1.2}
		  \vspace{-0.3cm}
		\begin{tabular}{XY|YYYY|YYYY||YYYY|YYYY}
            \hline
            \multicolumn{2}{l|}{Annotated Leixon Size} & \multicolumn{8}{c||}{5K all} & \multicolumn{8}{c}{1K unique}\\
            \hline
		    \multicolumn{2}{l|}{dataset}&\multicolumn{4}{c|}{MUSE} &\multicolumn{4}{c||}{VecMap}&\multicolumn{4}{c|}{MUSE} &\multicolumn{4}{c}{VecMap}\\
			\hline
			\multicolumn{2}{c|}{\multirow{2}{2cm}{Languages}}  & \multicolumn{2}{c}{EN-ES} & \multicolumn{2}{c|}{EN-FR} & \multicolumn{2}{c}{EN-DE} & \multicolumn{2}{c||}{EN-IT}& \multicolumn{2}{c}{EN-ES} & \multicolumn{2}{c|}{EN-FR} & \multicolumn{2}{c}{EN-DE} & \multicolumn{2}{c}{EN-IT}\\ 
			\cline{3-18}   
			& &$\rightarrow$&$\leftarrow$&$\rightarrow$&$\leftarrow$&$\rightarrow$&$\leftarrow$&$\rightarrow$&$\leftarrow$&$\rightarrow$&$\leftarrow$&$\rightarrow$&$\leftarrow$&$\rightarrow$&$\leftarrow$&$\rightarrow$&$\leftarrow$\\ \hline
			\multicolumn{10}{l||}{\textit{\textbf{Detailed Results of the Ablation to \name}}} & \multicolumn{8}{l}{\textit{\textbf{Detailed Results of the Ablation to \name}}}\\
			\arrayrulecolor{black} \cdashline{1-18}[1.5pt/2pt]
			\multirow{3}{2cm}{\name\textsf{- RCSLS}} & \textbf{best} & 84.6 & 87.1 & 85.5 & 85.3& 49.1 & 42.9 & 47.0 & 41.4 &83.8 &85.2 &84.0 &83.9 &48.2 &42.9 &45.7 &41.3 \\
    			 & \textbf{avg} &84.5&86.9&85.5&85.3& 48.4 & 42.5& 46.6 & 41.3 &83.8 &85.0 &83.9 &83.7 &47.8 &42.8 &45.3 &41.2 \\
    			 & \textbf{st} & 0.126 & 0.481 &  0.238 & 0.099 & 0.500   & 0.272&  0.425  & 0.178 &
    			 0.063 &0.144 &0.175 &0.136 &0.408 &0.071 &0.430 &0.096 \\ 
    		\arrayrulecolor{black} \cdashline{1-18}[1.5pt/2pt]
			\multirow{3}{2cm}{$\circleddash$ \POT} & \textbf{best}&84.0 &86.7 &84.1 &85.0&46.9 &42.0 &44.0 &39.6 & 83.5 &85.9 &83.0 &83.9 &42.5 &40.9 &37.1 &37.5 \\
    			 & \textbf{avg} &83.9 &86.6 &84.0 &84.8&46.6 &41.7 &43.7 &39.5&81.6 &84.8 &82.2 &83.4 &41.8 &40.2 &36.5 &35.7 \\
    			 & \textbf{st} &0.156 &0.099 &0.116 &0.158&0.330 &0.372 &0.282 &0.093&2.274 
&1.004 &1.372 &0.737 &1.175 &0.700 &0.835 &1.806  \\
			\arrayrulecolor{black} \cdashline{1-18}[1.5pt/2pt]
			\multirow{3}{2cm}{$\circleddash$ \BLU} & \textbf{best}&83.3 & 86.8 & 84.7 & 84.9 &47.7 &43.1 &45.6 &40.6&82.7 &83.9 &82.9 &82.3 &47.9 &43.0 &46.0 &40.1 \\
    			 & \textbf{avg} &83.2 &86.6 &84.4 &84.7& 47.4 &42.9 &45.4 &40.4&82.5 &83.7 &82.5 &82.2 &47.7 &42.6 &45.3 &39.7 \\
    			 & \textbf{st} &0.083 &0.201 &0.182 &0.180 & 0.247 &0.256 &0.219 &0.175&0.259 &0.197 &0.293 &0.115 &0.268 &0.314 &0.552 &0.343 \\
			\arrayrulecolor{black} \cdashline{1-18}[1.5pt/2pt]
			\multirow{3}{2cm}{$\circleddash$ \POT $\&$ $\circleddash$ \BLU} & \textbf{best}&82.9 &85.5 &83.3 &83.9&42.1 &37.0 &40.2 &34.9 &62.1 &63.2 &57.9 &60.5 &28.8 &22.6 &26.8 &24.1 \\
    			 & \textbf{avg} &82.5 &84.9 &83.0 &83.7 &41.7 &36.6 &39.7 &34.8&61.0 &62.4 &57.4 &59.5 &28.1 &22.4 &26.2 &23.7  \\
    			 & \textbf{st} &0.242 &0.396 &0.259 &0.205 &0.298 &0.234 &0.539 &0.075&0.756 &0.628 &0.412 &0.774 &0.476 &0.255 &0.410 &0.260 \\
			\arrayrulecolor{black} \cdashline{1-18}[1.5pt/2pt]	
			\multirow{3}{2cm}{$\circleddash$ \unsup $\&$ $\circleddash$ \POT} & \textbf{best}&84.4 &86.8 &85.5 &85.5&45.9 &40.6 &43.0 &39.5&83.4 &84.4 &83.3 &82.7 &42.1 &36.5 &38.6 &37.5 \\
    			 & \textbf{avg} &84.3 &86.5 &84.8 &85.1 &45.8 &40.1 &42.8 &39.0&81.7 &83.3 &80.5 &81.4 &40.3 &36.1 &38.2 &36.4 \\
    			 & \textbf{st} &0.191 &0.223 &0.500 &0.233 &0.144 &0.365 &0.360 &0.490&1.523 &1.346 &3.528 &1.506 &1.972 &0.465 &0.368 &0.998  \\
			\arrayrulecolor{black} \cdashline{1-18}[1.5pt/2pt]
			\multirow{3}{2cm}{$\circleddash$ \supe $\&$ $\circleddash$ \BLU} & \textbf{best}&82.4 &83.3 &82.6 &82.9& 48.1 &43.6 &45.5 &40.9&82.6 &83.9 &82.3 &83.0 &47.9 &43.0 &45.4 &40.4 \\
    			 & \textbf{avg} &82.3 &83.2 &82.5 &82.7& 47.7 &43.2 &45.3 &40.4&82.5 &83.8 &82.2 &82.9 &47.8 &42.8 &45.2 &39.7 \\
    			 & \textbf{st} &0.085 &0.115 &0.157 &0.347& 0.337 &0.530 &0.110 &0.388&0.115 &0.115 &0.135 &0.083 &0.040 &0.249 &0.148 &0.518 \\
			\hline	
			\multicolumn{10}{l||}{\textit{\textbf{Detailed Results of the Ablation to \DSS}}} & \multicolumn{8}{l}{\textit{\textbf{Detailed Results of the Ablation to \DSS}}}\\
			\arrayrulecolor{black} \cdashline{1-18}[1.5pt/2pt]
			\multirow{3}{2cm}{\DSS\textsf{- RCSLS} } & \textbf{best}& 83.9 & 87.0 & 84.6 & 85.5 & 51.2 & 45.1 & 48.3 & 42.5&83.1 &84.5 &82.4 &83.2 &48.8 &43.3 &47.1 &40.5 \\
    			 & \textbf{avg} &83.7 & 86.5 & 84.4 & 85.5 & 50.8 & 44.8 & 47.8 & 42.1&82.9 &83.8 &82.4 &83.0 &48.4 &43.0 &46.6 &40.1 \\
    			  &\textbf{st} &0.163 & 0.359 & 0.150 & 0.100 & 0.513  & 0.345 & 0.968 & 0.436 &0.139 &0.476 &0.063 &0.183 &0.409 &0.314 &0.451 &0.334 \\
    		\arrayrulecolor{black} \cdashline{1-18}[1.5pt/2pt]
			\multirow{3}{2cm}{$\circleddash$ \POT} & \textbf{best}&83.7 &86.2 &84.5 &85.5 &49.3 &43.8 &46.8 &40.9 &81.3 &83.5 &82.5 &82.4 &45.7 &41.8 &42.5 &37.1 \\
    			 & \textbf{avg} &83.5 &85.4 &84.4 &85.3 &49.1 &43.1 &46.5 &40.8 &81.1 &82.6 &82.4 &81.9 &45.1 &40.7 &41.7 &36.4 \\
    			 & \textbf{st} &0.162 &1.568 &0.087 &0.300 &0.205 &0.731 &0.446 &0.138 &0.348 &0.839 &0.135 &0.707 &0.816 &1.326 &1.222 &0.830 \\
			\arrayrulecolor{black} \cdashline{1-18}[1.5pt/2pt]
			\multirow{3}{2cm}{$\circleddash$ \BLU} & \textbf{best}&82.9 &85.7 &83.1 &84.4 &48.9 &44.0 &46.9 &40.2 & 82.1 &84.3 &82.5 &82.9 &48.5 &42.8 &45.3 &40.0 \\
    			 & \textbf{avg} &82.8 &85.4 &83.0 &84.3 &48.5 &43.7 &46.0 &39.9  & 81.9 &83.9 &82.2 &82.5 &48.0 &42.6 &44.8 &39.0 \\
    			 & \textbf{st} &0.083 &0.168 &0.083 &0.139 &0.391 &0.297 &0.768 &0.183 & 0.258 &0.252 &0.162 &0.301 &0.353 &0.199 &0.531 &0.746 \\
    		\hline
		\end{tabular}
	\caption{Detailed Experimental Results Ablation Study. We repeat the experiment on each language pair four times and report \textbf{best}, \textbf{avg}, \textbf{st} of the four results(\textbf{best}: the highest @1 accuracy. \textbf{avg}: the average accuracy which is reported in main body of this paper. \textbf{st}: the standard deviation.)}
	  \vspace{-2cm}
	\label{AT2}
\end{table*}

\end{document}